\documentclass[runningheads,a4paper]{llncs}
\usepackage{amssymb}
\usepackage{graphicx}
\usepackage{amsmath}
\usepackage{multirow}
\usepackage{color}
\usepackage{url}
\usepackage[normalem]{ulem}
\newcommand{\keywords}[1]{\par\addvspace\baselineskip
\noindent\keywordname\enspace\ignorespaces#1}
\newcommand{\RR}{\mathbb{R}}
\newcommand{\eps}{\epsilon}
\newcommand{\AUC}[1]{\mathrm{AUC}_\mathrm{#1}}

\begin{document}

\mainmatter  

\title{Layer-wise Relevance Propagation for Neural Networks with Local Renormalization Layers}

\titlerunning{LRP for Neural Networks with Local Renormalization Layers}

\author{Alexander Binder\inst{1} \and Gr\'egoire Montavon\inst{2} \and Sebastian Bach\inst{3} \and \mbox{Klaus-Robert M{\"u}ller\inst{2,4}} \and Wojciech Samek\inst{3}}
\authorrunning{Binder et al.} 
\institute{ISTD Pillar, Singapore University of Technology and Design \\
\and
Machine Learning Group, Technische Universit\"at Berlin\\
\and
Machine Learning Group, Fraunhofer Heinrich Hertz Institute\\
\and
Department of Brain and Cognitive Engineering, Korea University
}

\toctitle{Lecture Notes in Computer Science}
\tocauthor{Authors' Instructions}
\maketitle

\begin{abstract}
Layer-wise relevance propagation is a framework which allows to decompose the prediction of a deep neural network computed over a sample, e.g.~an image, down to relevance scores for the single input dimensions of the sample such as subpixels of an image. While this approach can be applied directly to generalized linear mappings, product type non-linearities are not covered. This paper proposes an approach to extend layer-wise relevance propagation to neural networks with local renormalization layers, which is a very common product-type non-linearity in convolutional neural networks. We evaluate the proposed method for local renormalization layers on the CIFAR-10, Imagenet and MIT Places datasets.
\keywords{Neural Networks, Image Classification, Interpretability}
\end{abstract}

\section{Introduction}

Artificial neural networks enjoy increasing popularity for image classification tasks. They have shown excellent performance in large scale competitions \cite{DBLP:conf/nips/KrizhevskySH12}. One reason is the ability to train neural networks with millions of training samples by parallelizing them on GPU hardware. This allows to use numbers of training samples which match the large number of parameters in deep neural networks.
However, understanding what region of the image is important for a classification decision, is still an open question for neural networks, as well as for many other non-linear models. The work of \cite{BacBinMonKlaMueSam15} proposed Layer-wise Relevance Propagation (LRP) as a solution for explaining what pixels of an image are relevant for reaching a classification decision. This was done for neural networks, bag of word models \cite{Csurka04,DBLP:journals/pami/SandeGS10}, and in a subsequent work \cite{bach-arxiv15}, for Fisher vectors.

This paper proposes an approach to extend LRP to neural networks with nonlinearities beyond the commonly used neural network formulation. One example of such nonlinearities are local renormalization layers which can not be handled by standard LRP \cite{BacBinMonKlaMueSam15}. The presented approach is based on first (or higher) order Taylor expansion. We consider a classification setup with real-valued outputs. A classifier $f$ is a mapping of an input space $f: X \rightarrow \RR$ such that $f(x)>0$ denotes the presence of the class.

\section{Layer-wise Relevance Propagation for Neural Networks}
\label{sec:lrp}

\begin{figure*}[t]
\centering
\includegraphics[width=0.16\textwidth]{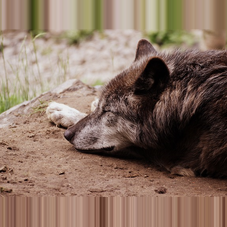}
\includegraphics[width=0.16\textwidth]{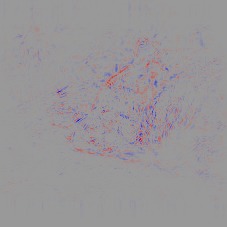}
\includegraphics[width=0.16\textwidth]{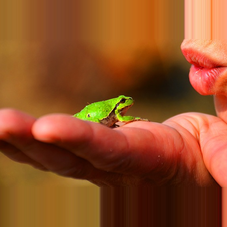}
\includegraphics[width=0.16\textwidth]{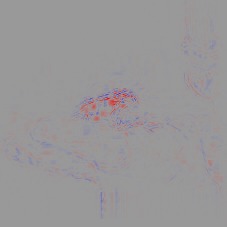}
\includegraphics[width=0.16\textwidth]{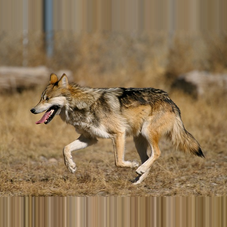}
\includegraphics[width=0.16\textwidth]{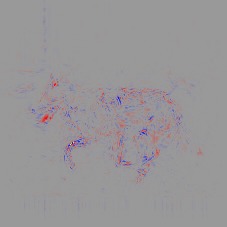}
\vskip -2mm
\caption{\label{fig:someheatmaps2} Pixel-wise decompositions for classes wolf, frog and wolf using a neural network pretrained for the 1000 classes of the ILSVRC challenge.}
\end{figure*}

In the following we consider neural networks consisting of layers of neurons. 
The output $x_j$ of a neuron $j$ is a non-linear activation function $g$ as given by
\begin{align}
x_j &=  g\Big(  {\textstyle \sum_i} w_{ij} x_i    +b \Big) 	\label{eq:neuron}
\end{align}
Given an image $x$ and a classifier $f$ the aim of layer-wise relevance propagation is to assign each pixel $p$ of $x$ a pixel-wise relevance score $R^{(1)}_p$ such that 
\begin{align}
f(x) \approx {\textstyle \sum_p} R^{(1)}_p \label{eq:pixwisedecompos}
\end{align}
\vskip -1mm
Pixels $p$ with $R^{(1)}_p<0$ contain evidence against the presence of a class, while $R^{(1)}_p>0$ is considered as evidence for the presence of a class. These pixel-wise relevance scores can be visualized as an image called \emph{heatmap} (see Fig.~\ref{fig:someheatmaps2} for examples). Obviously, many possible such decompositions exist which satisfy equation \ref{eq:pixwisedecompos}. The work of \cite{BacBinMonKlaMueSam15} yield pixel-wise decompositions which are consistent with evaluation measures \cite{DBLP:journals/corr/SamekBMBM15} and human intuition.

Assume that we know the relevance $R^{(l+1)}_j$ of a neuron $j$ at network layer $l+1$ for the classification decision $f(x)$, then we like to decompose this relevance into messages $R^{(l,l+1)}_{i \leftarrow j}$ sent to those neurons $i$ at the layer $l$ which provide inputs to neuron $j$ such that equation \ref{eq:lrc0} holds.
\begin{align}
  R^{(l+1)}_j=\sum_{i \in (l)} R^{(l,l+1)}_{i \leftarrow j} \label{eq:lrc0}
\end{align}
\vskip -1mm
We can then define the relevance of a neuron $i$ at layer $l$ by summing all messages from neurons at layer $l+1$ as in equation \ref{eq:lrc1}
\begin{align}
  R^{(l)}_i=\sum_{j \in (l+1)} R^{(l,l+1)}_{i \leftarrow j} \label{eq:lrc1}
\end{align}
\vskip -1mm
Equations \ref{eq:lrc0} and \ref{eq:lrc1} define the propagation of relevance from layer $l+1$ to layer $l$. The relevance of the output neuron at layer $M$ is $R^{(M)}_1=f(x)$. The pixel-wise scores are the resulting relevances of the input neurons $R^{(1)}_d$. 

The work in \cite{BacBinMonKlaMueSam15} established two formulas for computing the messages $R^{(l,l+1)}_{i \leftarrow j}$. The first formula called $\epsilon$-rule is given by
\begin{align}
R^{(l,l+1)}_{i \leftarrow j} & = \frac{ z_{ij} }{ z_j + \epsilon \cdot \mathrm{sign}(z_j) }R^{(l+1)}_j \label{eq:message1}
\end{align}
with $z_{ij} = (w_{ij} x_i )^p$ and $z_j = \sum_{k: w_{kj} \neq 0 } z_{kj}$.
The variable $\eps$ is a ``stabilizer'' term whose purpose is to avoid numerical degenerations when $z_j$ is close to zero, and which is chosen to be small. The second formula called $\beta$-rule is given by
\begin{align}
R^{(l,l+1)}_{i \leftarrow j} & = \Big((1+\beta)\frac{ z_{ij}^{+} }{ z_j^{+} }-\beta \frac{ z_{ij}^{-} }{ z_j^{-} } \Big)  R^{(l+1)}_j \label{eq:message2}
\end{align}
where the positive and negative weighted activations are treated separately. The variable $\beta$ controls how much inhibition is incorporated in the relevance redistribution. A fairly large value for $\beta$ (e.g. $\beta = 1$) leads to sharper heatmaps. In both formulas the message $R^{(l,l+1)}_{i \leftarrow j}$ has the following structure
\begin{align}
R^{(l,l+1)}_{i \leftarrow j} & = v_{ij} R^{(l+1)}_j \quad \mathrm{with} \quad {\textstyle \sum_i} v_{ij} = 1 
\end{align}
The meaningfulness of the resulting pixel-wise decomposition for the input layer $R_d^{(1)}$ comes from the fact that the terms $v_{ij}$ are derived from the weighted activations $w_{ij}x_i$ of the input neurons. Note that layer-wise relevance propagation does not use gradients in contrast to backpropagation during the training phase. For full details on layer-wise relevance propagation the reader is referred to \cite{BacBinMonKlaMueSam15}. 

\section{Extending LRP to local renormalization layers}
\label{sec:localrenormalization}

We consider a general neuron $j$ whose pooling and activation does not fit into the structure given by equation \ref{eq:neuron}, and consequently, intuition for a possible redistribution formula is lacking. In this paper we propose a strategy for such neurons, based on the Taylor expansion of its activation function. A Taylor-based approach was used in \cite{montavon2015explaining} for decomposing ReLU neurons by exploiting their local linearity. Here, we consider instead fully nonlinear neurons.

Suppose we can define for each neuron $i$ input to neuron $j$ a term $v_{ij}$ which is derived from its activation $x_i$ such that $\sum_i v_{ij}=1$. Then we can define a message $R^{(l,l+1)}_{i \leftarrow j} = v_{ij} R^{(l+1)}_j \label{eq:vijbasic}$. Such messages were used in equations \ref{eq:message1} and \ref{eq:message2} where the weighting $v_{ij}$ was chosen to depend on the weighted activations of neuron $i$: $v_{ij} = c \, (w_{ij} x_i )^p$ and $v_{ij} = c_1 z_{ij}^{+}+c_2 z_{ij}^{-}$, respectively. For differentiable neurons, such weighting can be obtained by performing a first order Taylor expansion. Let $x_j = g ( x_{h_1}, \ldots, x_{h_n})$ be a nonlinear activation function. Then, by Taylor expansion at some reference point $( \widetilde{x}_{h_1}, \ldots, \widetilde{x}_{h_n}  )$, we get
\begin{align}
x_j \approx g( \widetilde{x}_{h_1}, \ldots, \widetilde{x}_{h_n}  ) + \sum_{i \leftarrow j}\frac{\partial g} {\partial x_{h_i}}( \widetilde{x}_{h_1}, \ldots, \widetilde{x}_{h_n}  ) (x_{h_i}-\widetilde{x}_{h_i}).
\end{align}
Elements of the sum can be assigned to incoming neurons, and the zero-order term can be redistributed equally between them, leading to the decomposition
\begin{align}
\forall_{i \leftarrow j}:~ z_{ij}& =  \frac1n \,g( \widetilde{x}_{h_1}, \ldots, \widetilde{x}_{h_n}  ) + \frac{\partial g} {\partial x_{h_i}}( \widetilde{x}_{h_1}, \ldots, \widetilde{x}_{h_n}  ) (x_{h_i}-\widetilde{x}_{h_i}) \label{eq:vijdefinition}
\end{align}
of the neuron activation onto its input neurons. Local renormalization layers have been shown to improve the performance in deep neural networks \cite{DBLP:conf/nips/KrizhevskySH12}. Consider the local renormalization $y_k$ of a neuron $x_k$ by the set of its surrounding neurons $\{x_1, \ldots, x_n\}$ as 
\begin{align}
y_k(x_1, \ldots, x_n) & = \frac{x_k}{  \left(1+ b \sum_{i=1}^n x_i^2 \right)^c } \label{eq:lrn}
\end{align}
This interaction can be modeled by a layer in the network that has an activation function as given in equation \ref{eq:lrn}. Local renormalization layers represent a non-linearity which cannot be tackled exactly by LRP as introduced in \cite{BacBinMonKlaMueSam15}, however the strategy proposed above can be applied. 

One choice to be made is the point at which to perform the Taylor expansion. There are two apparent candidates, firstly the actual input to the renormalization layer 
$z_1=(x_1, \ldots, x_n)$ and, secondly, the input corresponding to the case when only the neuron $k$ fires which is to be normalized 
$z_2=(0, ... \ldots,0, x_k, 0, \ldots, 0)$. The partial derivative of $y$ at $z_2$ is zero for all variables $x_i$ with $i \neq k$ due to 
\begin{align}
\frac{\partial y_k}{\partial x_j}& =  \frac{\delta_{kj}}{  \left(1+ b \sum_{i=1}^n x_i^2 \right)^{c}  }-2bc\frac{x_k x_j}{  \left(1+ b \sum_{i=1}^n x_i^2 \right)^{c+1}  } 
\end{align}
This implies that the Taylor approximation has no off-diagonal contribution.
\begin{align}
 y_k(z_1) \approx y_k(z_2)+0=\frac{x_k}{(1+bx_k^2)^c} \label{eq:baseline2}
\end{align}
Therefore we apply the Taylor series around the point $z_1$:
\begin{align}
y_k(z_2) &\approx y_k(z_1) +\nabla y_k(z_1) \cdot (z_2-z_1)\\
\Rightarrow  y_k(z_1) &\approx y_k(z_2)  +\nabla y_k(z_1) \cdot (z_1-z_2) \\
\Rightarrow  y_k(z_1) &\approx \frac{x_k}{(1+bx_k^2)^c}-2bc \sum_{j:j \neq k} \frac{x_k x_j^2}{  \left(1+ b \sum_{i=1}^n x_i^2 \right)^{c+1}  } \label{eq:taylor_lrn_firstorder}
\end{align}
This weighting satisfies the following qualitative properties: for the neuron input $x_k$ which is to be normalized, the sign of the relevance is kept. For suppressing neighboring neurons $x_i$, $i \neq k$, the sign of the relevance can be flipped in line with their suppressing property. The absolute value of the relevance received by the suppressing neurons is proportional to the square of their input. In the limits $c \rightarrow 0$ and $b \rightarrow 0$, the local renormalization converges against the identity, and the approximation recovers the identity. 
A baseline to compare against is to treat the normalization as constant. In that case the weights $v_{ij}$ for the relevance propagation in equation \ref{eq:vijbasic} become a zero one vector, the relevance is propagated only to that neuron which is to be normalized: $v_{ij}=1$ if and only if $i$ is the neuron which is to be normalized by neuron $j$.

\section{Experiments}
\label{sec:experiments}
\begin{figure}[t]
\centering
\includegraphics[width=0.5\columnwidth]{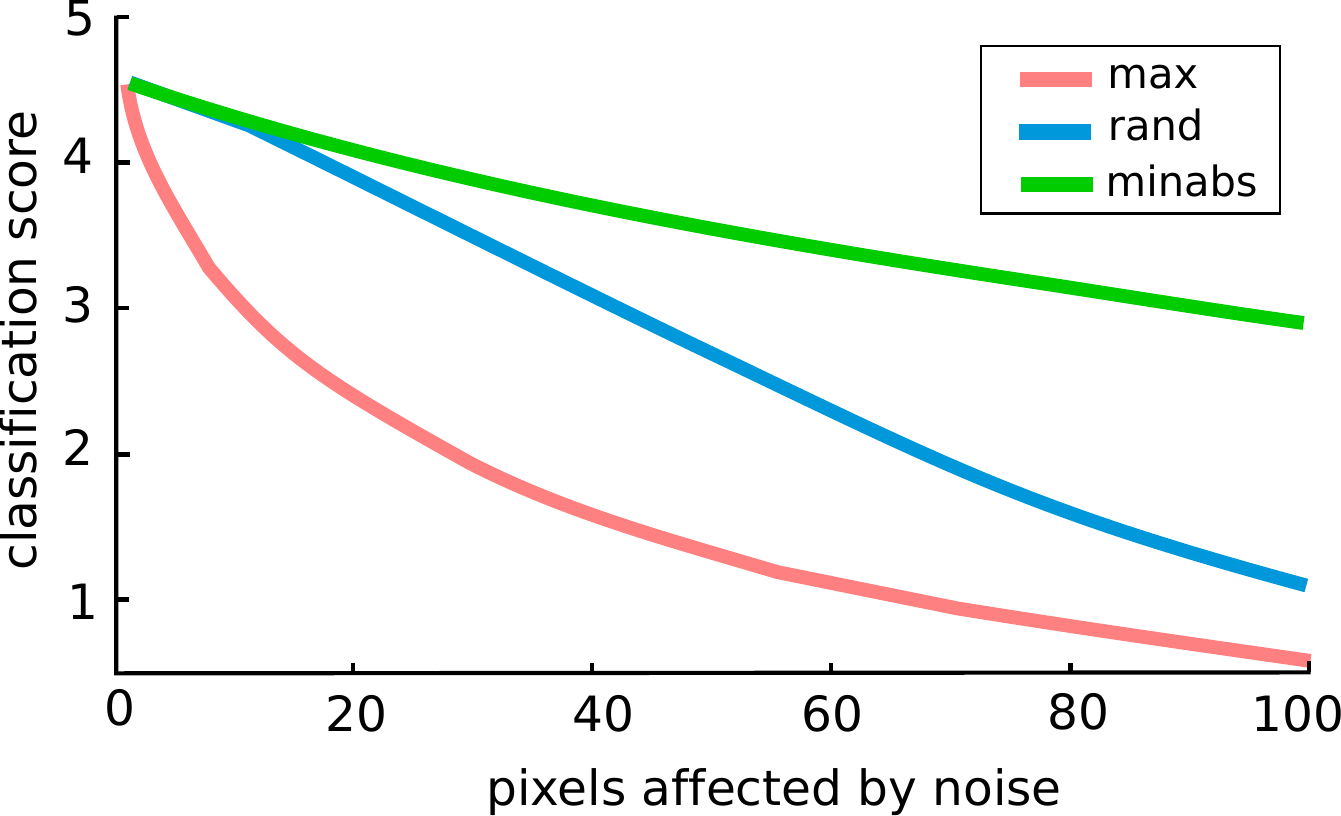}
\caption{\label{fig:threeflippingtypes}Decrease of classification score as pixels are sequentially replaced by random noise on the CIFAR-10 dataset. Red curve: pixels with highest pixel-wise scores are flipped first. Blue curve: pixels are flipped in random order. Green curve: least relevant pixels are flipped first. A similar comparison for Imagenet is found in \cite{DBLP:journals/corr/SamekBMBM15}.}
\end{figure}

We need to define a measure for meaningfulness and quality of a pixel-wise decomposition in order to evaluate the various strategies to compute it. Here we use an idea from \cite{DBLP:journals/corr/SamekBMBM15}: A pixel $p$ is considered highly relevant for the classification score $f(x)$ of the image $x$ if modifying it by assigning it a random RGB value $\tilde{x}(p)$, and classifying the modified image $\bar{x}_{p}= x \setminus\{x(p)\} \cup \{\tilde{x}(p)\}$ results in a strong decrease of the real-valued classification score $f(\bar{x}_{p})$. This idea can be extended by sequentially modifying pixels from the most relevant to the least relevant. The result is a graph of the prediction score $f(\bar{x})$ as a function of the number of modified pixels. An example for some sequences which will be explained below is shown in Fig.~\ref{fig:threeflippingtypes}. 
We can use these graphs to evaluate the meaningfulness of a pixel-wise decomposition.

In the first experiment we compare the measure when flipping highest-scoring pixels first, against flipping pixels in random order, and against flipping lowest scoring pixels first. If the classifier is able to identify pixels that are important for classification, then flipping highest scoring pixels first should result in the fastest decaying curve, while flipping lowest scoring pixels first should result in the slowest decrease. Fig.~\ref{fig:threeflippingtypes} tests this property on the CIFAR-10 dataset \cite{CIFAR-10} which consists of 50000 images of size $32\times 32$ drawn from 10 object classes. Scores are averaged over the 5000 images of the test set of CIFAR-10 for a classifier in which local renormalization layers are treated as the identity during computation of pixel-wise scores. Experiments corroborate that flipping highest scoring pixels first results in the fastest decrease of the prediction score on average over the test set. The decrease is sharper compared to random flipping, or flipping lowest scoring pixels first.

In a second experiment we compare which treatment of the local renormalization layer is best to identify those pixels that are most relevant for classifying an image. The two tested approaches for treating the local renormalization are (1) like it would be the identity, (2) by first order Taylor expansion as given by equation \ref{eq:taylor_lrn_firstorder}. These approaches are furthermore tested when used in conjunction with the two methods proposed by \cite{BacBinMonKlaMueSam15}, namely, the $\epsilon$-rule in equation \ref{eq:message1} with a fixed value of the numerical stabilizer $\epsilon$, and the $\beta$-rule shown in equation \ref{eq:message2}, with fixed $\beta$.

\begin{table}[t]
\centering
\begin{tabular}{ccc}\hline
~~~rule for basic layers~~~ & ~~~rule for normalization layers~~~ & ~~~AUC score~~~ \\ \hline
eq. \ref{eq:lrc1},\ref{eq:message1}, $\eps=0.01$ & identity & 37.10 \\
eq. \ref{eq:lrc1},\ref{eq:message1}, $\eps=0.01$ & first-order Taylor & 35.47\\
eq. \ref{eq:lrc1},\ref{eq:message2}, $\beta=1$ & identity & 56.13 \\
eq. \ref{eq:lrc1},\ref{eq:message2}, $\beta=1$ & first-order Taylor & 53.82 \\\hline
\end{tabular}
\vskip1em
\caption{\label{tab:expresults1}Comparison of different types of LRN layer treatments for two approaches of computing pixel-wise scores for CIFAR-10. Lower scores are better.}
\end{table}

We measure the quality of heatmaps by perturbing highest pixels first and computing the area under the curve (AUC). Lower AUC averaged over a large number of images indicates a better identification of pixel relevance by the heatmap. Results on CIFAR-10 are shown in Table \ref{tab:expresults1}. We observe that in all cases using first order Taylor in normalization layers improves the heatmap AUC score. This shows its effectiveness for dealing with non-linear neuron layers.

\begin{table}[t]
\centering
\begin{tabular}{c|ccccc}\hline
dataset & ~~methods~~ & $\Delta_{\epsilon=1}^{\epsilon=0.01}$ & \quad  $\Delta_{\epsilon=0.01}^{\epsilon=100}$ & \quad $\Delta_{\epsilon=1}^{\beta=1}$ & \quad $\Delta_{\beta=1}^{\beta=0}$  \\\hline
\multirow{2}{2cm}{\centering Imagenet}& identity & -21.29 & 2.75 & -42.61 & -49.07  \\
& Taylor & -12.29 & -41.75 & -34.44 & -50.76  \\\hline
\multirow{2}{2cm}{\centering MIT Places}& identity & -20.19 & 12.91 & -14.55 & -49.37  \\
& Taylor & -11.65 & -22.55 & -8.82 & -48.7  \\\hline
\end{tabular}
\vskip1em
\caption{\label{tab:order_imagenet}Comparison of different types of heatmap computations for Imagenet and MIT Places. We use the shortcut notation $\Delta_{a}^{b}$ for expressing $\AUC{a} - \AUC{b}$. Thus, a negative value indicates that the method produces better heatmaps with parameter $a$ than with parameter $b$. Note that $\eps$ refers to equations \ref{eq:lrc1} and \ref{eq:message1}; $\beta$ refers to eq.~\ref{eq:lrc1} and \ref{eq:message2}.}
\end{table}

We perform the same experiments also with Imagenet \cite{ILSVRC15} and MIT Places \cite{zhou2014learning} datasets, each time evaluating results for $5000$ images from their respective unlabeled test sets. Note that computing a heatmap requires only a predicted class label, not a ground truth. We evaluated results for the parameter settings $\beta=0$, $\beta=1$ in equation \ref{eq:message2} and $\eps=0.01$, $\eps=1$, $\eps=100$ in equation \ref{eq:message1}. Table \ref{tab:order_imagenet} shows the difference of AUC between variants of LRP, when using either the identity or the Taylor expansion for local renormalization layers. We observe the following ordering starting with the lowest (best) AUC: $\eps=1$, $\eps=0.01$, $\eps=100$, $\beta=1$, $\beta=0$. This order holds independent of whether we consider Imagenet or MIT places, when using Taylor for local renormalization layers. When using identity instead of Taylor, the order remain the same, except for $\eps=100$ and $\eps=0.01$ that are swapped. This is by itself an interesting result demonstrating that use of Taylor in the normalization layer does not disrupt the overall properties of relevance propagation techniques. For a comparison to other approaches such as heatmaps based on deconvolutions \cite{DBLP:conf/eccv/ZeilerF14}, or backpropagated gradients \cite{DBLP:journals/corr/SimonyanVZ13b} we refer to \cite{DBLP:journals/corr/SamekBMBM15}.

\begin{table}[t]
\centering
\begin{tabular}{c|cccccc}\hline
dataset &\quad  methods & \   $\eps=1 $ & \quad  $\eps=0.01 $ & \quad  $\eps=100$ &\quad  $\beta=1 $ &\quad  $\beta=0$  \\\hline
Imagenet & $\AUC{Taylor}-\AUC{identity}$ & \   -35.84 & \quad -26.84 &\quad  8.47 & \quad 0.29 & \quad 1.98 \\
MIT Places & $\AUC{Taylor}-\AUC{identity}$ & \  -33.13 & \quad -24.59 & \quad 5.34 &\quad  -0.39 & \quad -1.06 \\\hline
\end{tabular}
\vskip1em
\caption{\label{tab:lrn_nolrn_imagenet} Impact of using the Taylor method in various settings. Negative value indicates that using the Taylor expansion for the local renormalization is better in AUC terms (i.e.\ heatmaps are more representative of the importance of each pixel).}
\end{table}

\begin{figure}[t]
\centering \small
\rotatebox{90}{~~~~~~~image}
\includegraphics[width=0.2\columnwidth]{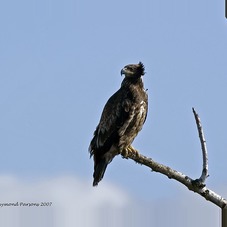} 
\includegraphics[width=0.2\columnwidth]{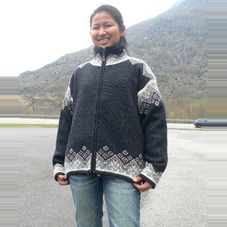}
\includegraphics[width=0.2\columnwidth]{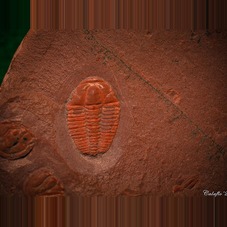} 
\includegraphics[width=0.2\columnwidth]{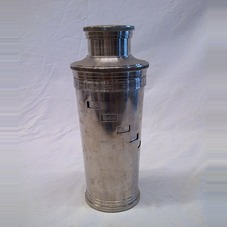}\\[1mm]
\rotatebox{90}{~~identity, $\epsilon=1$}
\includegraphics[width=0.2\columnwidth]{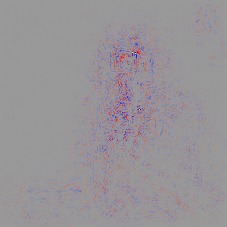} 
\includegraphics[width=0.2\columnwidth]{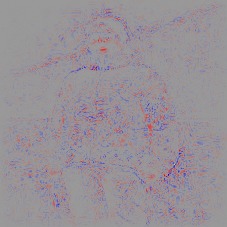} 
\includegraphics[width=0.2\columnwidth]{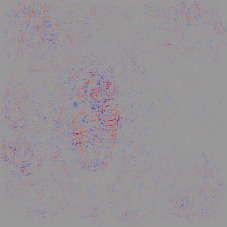} 
\includegraphics[width=0.2\columnwidth]{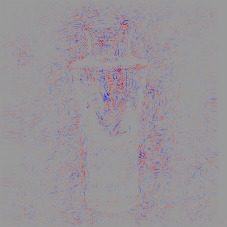}\\[1mm]
\rotatebox{90}{~~~Taylor, $\epsilon=1$}
\includegraphics[width=0.2\columnwidth]{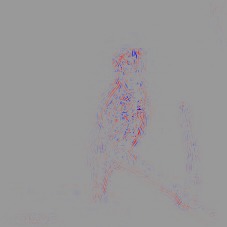} 
\includegraphics[width=0.2\columnwidth]{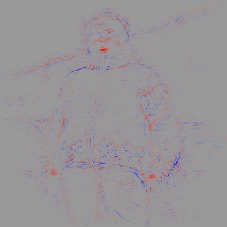} 
\includegraphics[width=0.2\columnwidth]{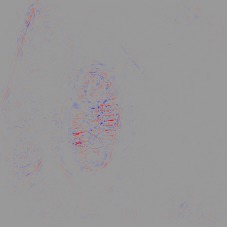} 
\includegraphics[width=0.2\columnwidth]{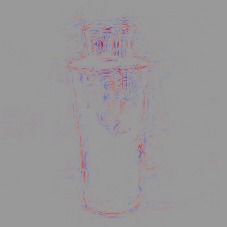}\\[1mm]
\rotatebox{90}{~~identity, $\beta=0$}
\includegraphics[width=0.2\columnwidth]{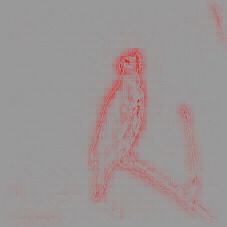} 
\includegraphics[width=0.2\columnwidth]{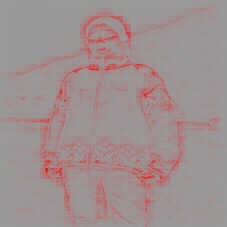} 
\includegraphics[width=0.2\columnwidth]{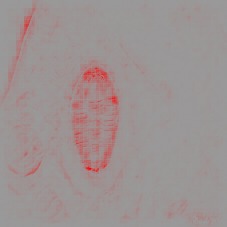} 
\includegraphics[width=0.2\columnwidth]{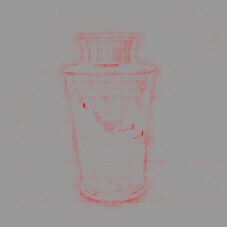}\\[1mm]
\caption{\label{fig:examples_lrn_vs_nolrn}Top row shows original unwarped image. Remaining rows show heatmaps produced by various parameters of the LRP method.}
\end{figure}

Table \ref{tab:lrn_nolrn_imagenet} shows the difference of AUC between Taylor and identity for local renormalization layers, for various choices of datasets and LRP parameters. We observe that for the parameters with best AUC ($\eps=1$ and $\eps=0.01$), using Taylor expansion for representing local renormalization layers further improves the AUC scores. For the remaining choices the results are on par or slightly worse. This is consistent with the interpretation of large values of $\eps$ as smoothing out small contributions. It is also consistent with the observation that $\beta=1$ and $\beta=0$ yield both smooth heatmaps in general. Heatmaps for some parameters of interest are shown in Figure \ref{fig:examples_lrn_vs_nolrn}. Taylor with $\epsilon=1$ has both high pixel selectivity and low noise, which in agreement with its measured superiority in the quantitative experiments.

\section{Conclusion}

We have presented an extension of layer-wise relevance propagation (LRP) based on first-order Taylor expansions for product-type nonlinearities. Such nonlinearities occur in the local renormalization layers of deep convolutional neural networks. The proposed extension is evaluated on three popular datasets and it is shown to clearly outperform the original LRP method. In future work we will investigate the potential gain of using higher order Taylor expansions, and apply the method to a larger class of neural network layers.


\bibliographystyle{plain}
\bibliography{snippet}
\end{document}